\documentclass[10pt,twocolumn,letterpaper]{article}

\usepackage{iccv}
\usepackage{times}
\usepackage{epsfig}
\usepackage{graphicx}
\usepackage{amsmath}
\usepackage{amssymb}

\usepackage{xcolor}
\usepackage{colortbl}
\usepackage{arydshln}

\usepackage{amssymb}
\usepackage{pifont}
\newcommand{\cmark}{\ding{51}}%
\newcommand{\xmark}{\ding{55}}%
\definecolor{ForestGreen}{RGB}{34,139,34}
\usepackage[accsupp]{axessibility}
\usepackage{balance}

\usepackage[pagebackref=true,breaklinks=true,letterpaper=true,colorlinks,bookmarks=false]{hyperref}

 \iccvfinalcopy 

\def\iccvPaperID{10478} 

\ificcvfinal\fi

\begin{document}

\title{Alignment-free HDR Deghosting with Semantics Consistent Transformer}

\author{Steven Tel$^{1,5}$\thanks{Both authors contributed equally to this research.}  \quad Zongwei Wu$^{2,5}$\footnotemark[1]
\quad Yulun Zhang$^{3}$ \thanks{Corresponding Author: Yulun Zhang} \quad Barthélémy Heyrman$^{1}$  \\ Cédric Demonceaux$^{4,5}$ \quad Radu Timofte$^{2}$ \quad Dominique Ginhac$^{5}$
\\
\small ${}^{1}$ University of Burgundy, ImViA \quad 
\small ${}^{2}$ Computer Vision Lab, CAIDAS \& IFI, University of W\"{u}rzburg
\\ 
\small ${}^{3}$ CVL, ETH Z\"{u}rich \quad 
\small ${}^{4}$ University of Lorraine, CNRS, Inria, Loria \quad 
\small ${}^{5}$ University of Burgundy, CNRS, ICB   \\
\footnotesize  \{steven.tel; barthelemy.heyrman; cedric.demonceaux; dginhac\}@u-bourgogne.fr,  \{zongwei.wu; radu.timofte\}@uni-wuerzburg.de,  yulun100@gmail.com
}

\maketitle
\ificcvfinal\thispagestyle{empty}\fi

\begin{abstract}
High dynamic range (HDR) imaging aims to retrieve information from multiple low-dynamic range inputs to generate realistic output. The essence is to leverage the contextual information, including both dynamic and static semantics, for better image generation. Existing methods often focus on the spatial misalignment across input frames caused by the foreground and/or camera motion. However, there is no research on jointly leveraging the dynamic and static context in a simultaneous manner.  To delve into this problem, we propose a novel alignment-free network with a Semantics Consistent Transformer (SCTNet) with both spatial and channel attention modules in the network. The spatial attention aims to deal with the intra-image correlation to model the dynamic motion, while the channel attention enables the inter-image intertwining to enhance the semantic consistency across frames. Aside from this, we introduce a novel realistic HDR dataset with more variations in foreground objects, environmental factors, and larger motions. Extensive comparisons on both conventional datasets and ours validate the effectiveness of our method, achieving the best trade-off on the performance and the computational cost. The source code and dataset are available at \url{https://steven-tel.github.io/sctnet/}.
\end{abstract}

\section{Introduction}
\label{sec:introduction}

While dealing with the full range of illumination in natural scenes, traditional cameras fail to provide images with reasonable quality and result in the loss of information in under-exposed and over-exposed areas. Some specialized hardware devices~\cite{855857,1467255} have been proposed to produce HDR images directly. Unfortunately, they are usually too expensive to be widely adopted. To this end, computational High Dynamic Range (HDR) imaging has recently attracted great research attention, aiming to merge multiple Low Dynamic Range (LDR) images with different exposure times to produce an HDR image with increased dynamic range and heightened realism.

Given well-aligned LDR images, traditional imaging methods~\cite{devebec,PatchBasedHDR,Hu,Kalantari2017,deephdr} can produce faithful results. Nevertheless, a perfect alignment is always hard to achieve. In practice, it is very common to have dynamic objects in the foreground and/or with ego motion, resulting in undesirable misalignment between the LDR inputs. Moreover, the movement across frames with different exposure values (EV) may lead to missing content in the over-/under-exposed regions. Therefore, a method that can efficiently leverage information from multiple exposures, while being robust to the misalignment is highly desired.

\begin{figure}[t]
\centering
\includegraphics[width=\linewidth,keepaspectratio]{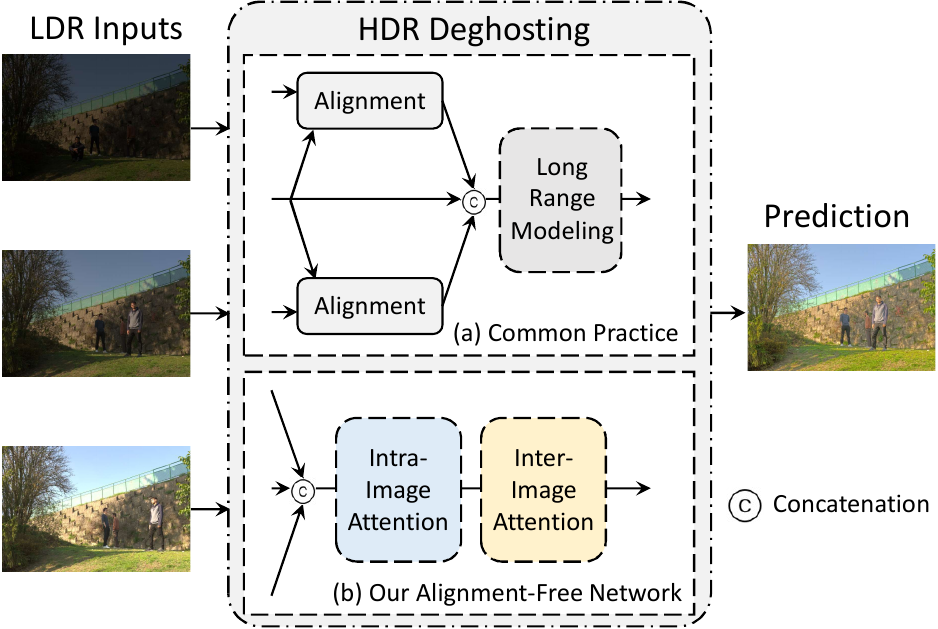}
\vspace{-4mm}
\caption{Motivation of our alignment-free network. Most existing methods \cite{hdrtransformer,tan2021deep,huang2022hdr,pu2020robust,yan2022lightweight} calibrate the LDR inputs at shallow latent space vis-à-vis the dynamic motion. Differently, we propose an alignment-free network that jointly models the dynamic and static context across frames through attention blocks.}
\label{fig:1}
\vspace{-2mm}
\end{figure}

To tackle the misalignment issue due to the moving object, multiple methods~\cite{Bogoni,kang} propose to compute the optical flow to model the foreground motion. Several recent methods~\cite{Kalantari2017,mostafavi2021learning,catley2022flexhdr} further integrate optical flow into a learning-based method to align motion between the input images. However, due to the poor generalization of optical flow methods on multiple exposures, these methods fail to produce reliable results. Other methods \cite{ahdrnet,hdrtransformer} compute the affinity matrices between different images with the help of attention modules as shown in Figure~\ref{fig:1}. Once the multi-EV features are calibrated, they are further merged and fed into deep convolutional neural networks (CNNs) such as \cite{zhang2018residual}. More recently, Niu~\etal~\cite{hdrgan} leverages generative adversarial network (GAN) for HDR imaging to deal with ghost artifacts. \cite{huang2022hdr} further merges GAN with an attention module to improve the model efficiency. However, these methods are based on local convolutions, limiting their ability to model and benefit from long-range dependencies. A recent method~\cite{hdrtransformer} replaces the deep convolutional network with a transformer and proposes a hybrid CNN-transformer architecture.  In general, existing methods \cite{hdrtransformer,ahdrnet,huang2022hdr,pu2020robust,yan2022lightweight}often follow the common practice of first computing the cross-image alignment with convolutional attention. Then, the features are further refined with contextualized awareness. Despite the plausible performance, these methods are heavily based on the alignment module for multi-EV feature calibration. Moreover, the SOTA performance is achieved with heavy computational cost, which is undesirable for real-time applications.

We observe that previous works \cite{pu2020robust,yan2022lightweight,catley2022flexhdr,ahdrnet,tan2021deep} often focus on dealing with the motion changes across different images. However, motion changes are highly correlated, not decoupled, with the static contexts, \ie, the semantic information of the holistic environment during imaging. Inspired by this observation, in this paper, we propose SCTNet, a novel Semantics Consistent Transformer network, which consists of two parts: intra-image spatial feature modeling and inter-EV semantic intertwining. Specifically, instead of computing the attention at a shallow stage, as shown in Figure~\ref{fig:1}, we directly investigate the transformer network to explore the spatial correlation between the merged multi-EV inputs. This spatial feature modeling operation aims to not only compute the contextualized awareness within a single image but also explore the spatial correlation across LDR inputs. However, directly aggregating the multi-input features is challenging since the different exposures may have different spatial distributions. Thus, we complement the spatial attention block with our new semantic-consistent cross-frame attention block. These two attention blocks work in a consecutive manner to make sure that both the foreground motion and the static context are well-learned and merged simultaneously in an end-to-end manner.

Moreover, to fully benefit from the popular non-local attention mechanism, we introduce a novel dataset for HDR deghosting that adheres to the standard acquisition approach put forth by \cite{Kalantari2017}. Our dataset is collected to fulfill the demand for larger training and evaluation data required to implement vision transformers in HDR deghosting tasks. The dataset comprises diverse global and local motion and illumination ranges to generate more versatile and all-encompassing HDR deghosting solutions. To make algorithms more robust, we have expanded our dataset to include a wider range of scenes, including twilight and night scenes, in contrast to the conventional dataset~\cite{Kalantari2017} with only day scenes.  To summarize, our contributions are threefold:
\vspace{-2mm}
\begin{itemize}
\setlength{\itemsep}{0pt}
\setlength{\parsep}{0pt}
\setlength{\parskip}{0pt}
    \item We propose a unified framework, SCTNet, that jointly models the foreground motion and static semantic context simultaneously, both of which are essential for HDR deghosting.
    \item We introduce a realistic supervised HDR deghosting dataset presenting a richer variety of scenes with more variation in the types of motion and more environmental factors than existing benchmark datasets.
    \item We demonstrate the effectiveness of our approach through extensive experiments on both the Kalantari dataset~\cite{Kalantari2017} as well as our proposed dataset. Our SCTNet performs favorably against other state-of-the-art techniques by reaching the best trade-off between performance and computational cost.
\end{itemize}

\section{Related works}
\label{sec:related_works}

\noindent \textbf{HDR Deghosting Methods:}\label{ssc:HDR_deghosting_methods}
Existing HDR deghosting algorithms can be categorized into two groups: Patch Match-Based and Deep Learning methods.

\noindent \textbf{\textit{Patch Match-Based Methods}}: One category involves pixel rejection techniques. These methods focus on globally aligning LDR images and then discarding misaligned pixels. Approaches include error map creation based on color differences by Grosch~\etal~\cite{Grosch2006FastAR}, motion area identification using threshold bitmaps by Pece~\etal~\cite{Pece}, and weighted intensity variance analysis by Jacobs~\etal~\cite{jacobs}. Other methods propose gradient-domain weight maps by Zhang~\etal~\cite{zhang} and probability maps by Khan~\etal~\cite{Khan2006GhostRI}. However, such methods often lead to suboptimal HDR results due to pixel rejection causing information loss.

Another set of methods focuses on motion registration. These involve aligning non-reference LDR images with the reference LDR image prior to merging. Techniques include optical flow-based motion vector prediction by Bogoni~\cite{Bogoni}, luminance domain transformation using exposure time and optical flow-based motion compensation by Kang~\etal~\cite{kang}, and optical flow-based registration followed by HDR reconstruction by Zimmer~\etal~\cite{Zimmer}. Additionally, Sen~\etal~\cite{PatchBasedHDR} introduced a patch-based energy minimization for simultaneous alignment and HDR reconstruction. Hu~\etal~\cite{Hu} enhanced image alignment through brightness and gradient consistency optimization.

\noindent \textbf{\textit{Deep Learning Solutions}}: To address occlusions in patch-based methods, convolutional neural networks (CNNs) gained prominence due to their ability to handle such challenges more effectively. Kalantari~\etal~\cite{Kalantari2017} pioneered the integration of a sequential neural network into the alignment-before-merging pipeline using optical flow. Wu~\etal~\cite{deephdr} adopted a direct domain translation network for ghost-free HDR generation. Further solutions utilized attention mechanisms to better handle long-range ghosting effects. Yan~\etal~\cite{ahdrnet} employed spatial attention, while Yan~\etal~\cite{NHDRRNet} used a non-local network for merging. Niu~\etal~\cite{hdrgan} introduced a GAN-based training scheme for high-quality ghost-free HDR generation.

Transformers have emerged as a promising solution due to their self-attention mechanism. Song~\etal~\cite{transhdr} proposed a transformer-based selective HDR reconstruction network, and Liu~\etal~\cite{hdrtransformer} introduced a hierarchical network combining Transformers and convolutional channel attention to capture global and local dependencies, respectively.

\noindent\textbf{HDR Deghosting Datasets:}
\label{ssc:HDR_deghosting_datasets}
The first dataset with HDR labels for training supervised HDR deghosting 
 networks have been proposed by Kalantari~\etal~\cite{Kalantari2017}. Each scene is composed of a set of three LDR input images of the same scene separated by two or three exposure values (EV). Motion in each input image is achieved using a human subject moving in the foreground of a static scene. For each scene, the corresponding HDR ground truth is generated from a static set of three input images (in which the 
 subject is requested to stay still during the acquisitions), using a triangle weighting function similar to that proposed by Debevec~\etal~\cite{devebec}. The middle exposure input from the static bracket is used as
 the middle input of the sample. This method enables the creation of realistic HDR ground truth. However, acquiring a large number of samples can be a tedious and time-consuming process, as it involves multiple iterations to ensure that scenes without any motion are selected for ground truth generation. For example,  Prahakbar~\etal~ \cite{Prabhakar2020TowardsPA} utilized this method to create a dataset. Nonetheless, the resulting ground truth images often contain ghosting artifacts.
 
Another approach for generating a three-input HDR deghosting dataset was introduced by the NTIRE HDR challenge~\cite{ntire21}. To create a more diverse dataset, the HDR ground-truth images were obtained from the work of Froelich~\etal~\cite{Froelich}, who captured a comprehensive collection of HDR videos using a professional two-camera rig with a semitransparent mirror for the purpose of evaluating HDR displays. Since these images do not have the required LDR input images, the corresponding LDR counterparts were generated synthetically using image formation models that include noise sources.


\begin{table}[t]
\begin{center}
\small	
\setlength\tabcolsep{4pt}
\renewcommand{\arraystretch}{1.0}
\caption{Comparison against existing benchmarks. Our proposed benchmark is more complete, providing more realistic and challenging samples for HDR deghosting.}
\begin{tabular}{c c c c c c }
\hline

\hline

\hline
Dataset & P.A & R.W & U.G.T & V.L.C. & I.C.\\
\hline
Kalantari~\etal~\cite{Kalantari2017} & \cmark & \cmark & \cmark & \xmark  & \cmark\\
Prabhakar~\etal~\cite{Prabhakar2020TowardsPA} & \xmark & \cmark & \xmark & \xmark & \cmark\\
NTIRE21~\cite{ntire21} & \cmark & \xmark & \cmark & \cmark & \xmark\\ 
\hdashline
\textbf{Ours} & \cmark & \cmark &  \cmark & \cmark & \cmark\\
\hline

\hline
\multicolumn{6}{l}{Attribute Description:}\\
P.A. & \multicolumn{5}{l}{Publicly Accessible} \\
R.W. & \multicolumn{5}{l}{Real World dataset} \\
U.G.T. & \multicolumn{5}{l}{Unghosted Ground Truth} \\
V.L.C. & \multicolumn{5}{l}{Variety of Light Conditions} \\
I.C. & \multicolumn{5}{l}{Information Consistency} \\

\hline

\hline

\hline
\vspace{-10mm}
\end{tabular}
\end{center}
\label{tab:dataset}
\end{table}

\section{Our Dataset}
\label{sec:our_benchmark}

\begin{figure}[h]
\begin{center}
   \includegraphics[width=1\linewidth]{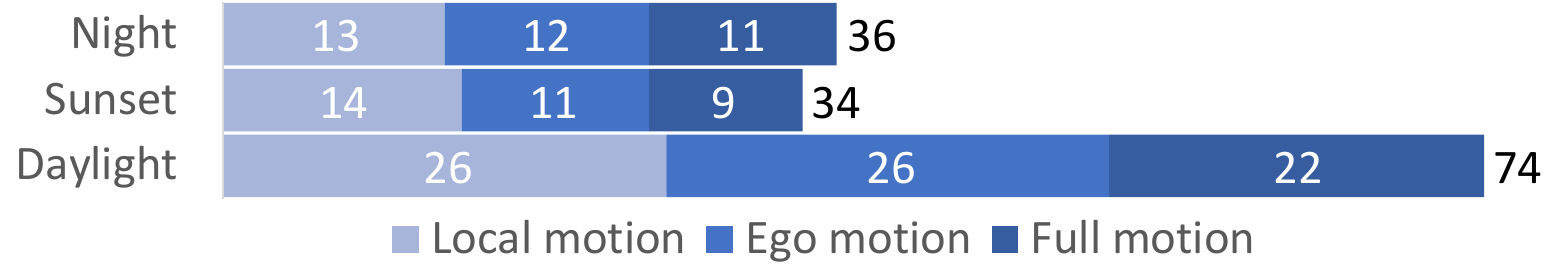}
\vspace{-5mm}
   \caption{Scene and motion distributions of our proposed dataset.}
   \vspace{-4mm}
\label{fig:data_distribution}
\end{center}
\end{figure}

\begin{figure*}
\begin{center}
\includegraphics[width=\linewidth,keepaspectratio]{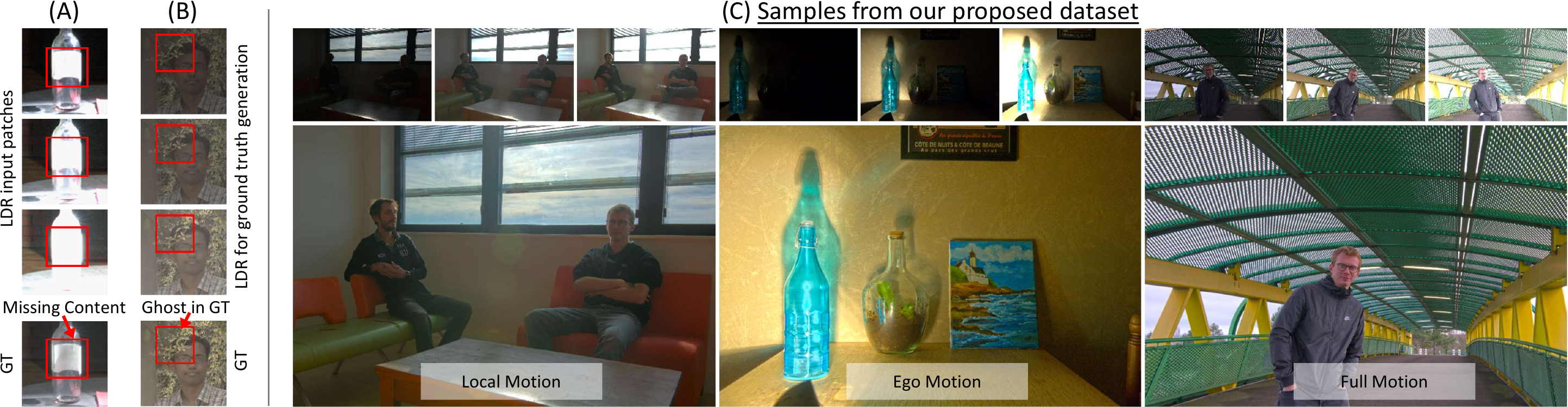}
\end{center}
\vspace{-2mm}
   \caption{Comparison with existing datasets. (A) NTIRE dataset~\cite{ntire21} may contain content in the ground truth (GT) images while not being available from the LDR images. (B) Prabhakar~\cite{Prabhakar2020TowardsPA} dataset generates the GT image with ghosts, which is mainly due to the background motion in the LDR images. (C) Our dataset contains challenging samples with different kinds of motion. From left to right: local motion where only people are moving, ego motion where only the camera is moving, and full motion where both foreground and camera are moving. Please zoom in for more details.}
   \vspace{-3mm}

\label{fig:motion_dataset}
\end{figure*}

\subsection{Comparison with state-of-the-art HDR datasets}

In Table~\ref{tab:dataset}, we highlight the strengths of our dataset over existing HDR datasets. Specifically, compared to the widely used Kalantari~\cite{Kalantari2017} dataset, our dataset offers a wider variety of scenes including day, twilight, and night scenarios, while Kalantari only includes scenes captured during the day. Moreover, 
the variety of movements in our samples is richer because our dataset includes not only the human subjects as foreground motion but also ego motion and combined human subjects/camera motion. Besides, our dataset contains more scene samples for both training and testing, \ie, 108/36 training/testing samples compared to 74/15 in the Kalantari dataset. It is also noteworthy that the scene variation of the Kalantary dataset is limited to the daytime and only 3 among the 15 testing samples contain extremely light camera motion, restricting its evaluative capacity. Conversely, our dataset addresses these limitations with a more \textbf{balanced scene and motion distribution}, as shown in Figure \ref{fig:data_distribution}. We collected samples from diverse light conditions (day, sunset, and night) and encompassed \textbf{richer motion types} (local, ego, and full motions) in equal proportions, for both train/test samples. This diverse motion range facilitates a comprehensive assessment of HDR methods across challenging scenes.

During recent challenges~\cite{ntire21}, other HDR samples have been released. However, the associated LDR images are generated from HDR images with different sources of synthetic noise, resulting in non-faithful inputs. As shown in Figure~\ref{fig:motion_dataset}, there may exist information in the ground truth (GT) images, while being not retrievable from the input LDR images. 
Finally, the recent Prabhakar dataset ~\cite{Prabhakar2020TowardsPA} claims to have more than 500 samples, but, until now, only 32 of them are publicly available. Moreover, as shown in Figure~\ref{fig:motion_dataset}, the GT images from the Prabhakar dataset may contain ghosts, which severely limits its application. 

In contrast to these state-of-the-art datasets, our dataset is more realistically complete because it was built from high-quality LDR inputs and artifact-free HDR ground truth. Thanks to the large variation in both motion types and lighting conditions, our dataset can be naturally used for analyzing the generalization capability across different scenarios.

\subsection{Acquisition process}

Our objective is to produce a set of LDR images with controlled motion and the GT HDR image.
Before any acquisition, we first select the exposure time of the reference image to be certain to get the most information on the scene. For the acquisition process, we use bracketings of 9 images spaced by  $\pm$ i-EV difference where $i\in{1, 2, 3, 4}$ around the reference exposure. Using 9 exposure values guarantees to cover the full dynamic range of a scene. A tripod-mounted camera is used to perform sequentially the 9 acquisitions of a static scene, for which we check that no parasitic motion exists. All the images are captured in raw format with a resolution of 4256 x 2832 pixels using a Nikon D700 camera.
Motion in each scene is controlled by a stop motion technique including 3 stops of 9 images.
In order to provide an enhanced dataset including greater opportunities to accurately assess the deghosting capabilities of current and future HDR solutions, we propose three different kinds of motions for each scene, as illustrated in Figure~\ref{fig:motion_dataset}: \\
 \textbf{Local motion:} The scene is static. Movement between frames is introduced by people or objects in the foreground; \\
 \textbf{Ego motion:} Background and foreground of the scene are static, the deghosting challenge is introduced in all of the pixels of the scene due to the camera ego motion; \\
 \textbf{Full motion:} Composed of both ego and local motions, these scenarios are the most challenging.

\subsection{Sample generation}

\begin{figure*}[t]
\begin{center}
\includegraphics[width=\linewidth,keepaspectratio]{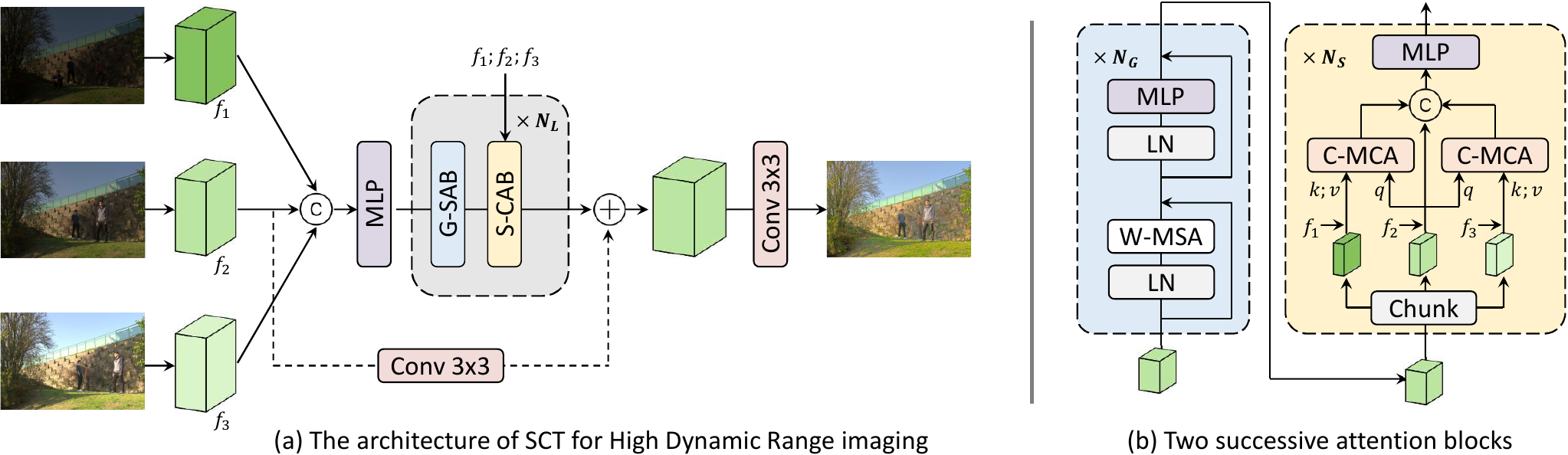}
\end{center}
\vspace{-4mm}
   \caption{The architecture of our proposed SCTNet for HDR deghosting. (b) The inner structure of two successive attention blocks G-SAB and S-CAB. The G-SAB leverages the spatial awareness to deal with the intra-image correlation to model the dynamic motion, while our S-CAB block attends to the channel direction and enables the inter-image intertwining to enhance the semantic consistency across frames.}
      \vspace{-3mm}
\label{fig:network_arch}
\end{figure*}

 Since there is no motion between the 9 images of each stop, the method proposed by Debevec~\cite{devebec} is applied to produce a ground truth HDR image (named $HDR_9$) for the second stop by merging the 9 input images. 
 Following the common HDR practice with only 3 LDR inputs, we also have to select the 3 best images from the 9 available candidates in each stop, while maintaining the most information possible. For this purpose, the reference image and the images with $\pm$ i-EV difference, $i\in{1, 2, 3, 4}$, are selected to produce the different $HDR_{3_i}$ images.
 In order to preserve the informative features from each input image, the triangle functions proposed by \cite{Kalantari2017} are used as blending weights $w_n$ to generate the ground truth HDR images:
 
\begin{equation}\label{eq:blendingfunction}
\hat{H}(p) = \frac{\sum_{n=1}^{3} w_n(p)H_n(p)}{\sum_{n=1}^{3} w_n(p)},
\end{equation}
where the resulting HDR image $\hat{H}$ at the pixel $p$ is the sum of the projected HDR inputs $H_n$ obtained following the gamma projection function described in Eq.~\ref{eq:hi}. 

By computing the HDR-VDP2~\cite{hdrvdp2} metrics between $HDR_9$ and each of the  $HDR_{3_i}$ images, we choose the
$HDR_{3_i}$ producing the best score as the ground truth HDR image of the sample scene, leading to 200 samples. To ensure high-quality of our dataset, we invited five viewers for quality control and rejected the unfaithful results. Our final benchmark contains 108/36 training/testing samples.


\section{Our Method}\label{sec:our_SCTNet}

\subsection{Problem Setup}
\label{ssc:problem_setup}

We define $L_k$ as the input sequence of $k$ LDR images from a dynamic scene with different exposures, HDR imaging aims at reconstructing an HDR image $\hat{H}$ aligned to a selected reference image $L_R$. Following the common practice, we chose $k=3$ as the number of input images, and select the middle exposure $L_2$ input as the reference input. To increase the robustness of our solution to exposure differences between input frames, we compute the respective projection of each LDR input frame into the HDR domain using the gamma encoding function, as described in Eq.~\ref{eq:hi}:
\begin{equation}\label{eq:hi}
H_{i} = \frac{L_{i}^\gamma}{t_{i}}, \quad \gamma=2.2 ,
\end{equation}
where $t_i$ is the exposure time corresponding to the image $i$, $H_i \in \mathbb{R} ^{3 \times H \times W}$ is the gamma-projected input of the corresponding image $I_i$. By concatenating the LDR images $L_i$ with $H_i$, we obtain $I_i \in  \mathbb{R} ^{6 \times H \times W}$ for each EV input. 

\subsection{Network Architecture}
\label{ssc:network_architecture}

\begin{table*}[t]
\small
\begin{center}
\caption{Quantitative comparison with  state-of-the-art methods on the Kalantari~\cite{Kalantari2017} testing samples. $l$-PSNR and $l$-SSIM are computed in the linear domain while $\mu$-PSNR and $\mu$-SSIM are computed after $\mu$-law tone mapping (Eq.~\ref{eq:mulaw}). PU-PSNR and PU-SSIM are computed by applying the encoding function proposed in \cite{pu21}.}
\label{table:quant}
\setlength\tabcolsep{10pt}
\begin{tabular*}{\linewidth}{@{\extracolsep{\fill}} l|rrrrrrr }
\hline

\hline

\hline
Method & $\mu$-PSNR & PU-PSNR & $l$-PSNR& $\mu$-SSIM & PU-SSIM & $l$-SSIM & HDR-VDP2\\
\noalign{\smallskip}
\hline
\noalign{\smallskip}
Hu \etal\cite{Hu} & 35.79 & 34.96 & 30.76  & 0.9717 & 0.9615 & 0.9503  & 57.05 \\
Sen \etal\cite{PatchBasedHDR} & 40.80 & 40.47 & 38.11  & 0.9808 & 0.9775 & 0.9721 & 59.38 \\
DHDRNet\cite{Kalantari2017} & 42.67 & 41.83 & 41.23  & 0.9888 & 0.9832 & 0.9846  & 65.05 \\
DeepHDR\cite{deephdr} & 41.65 & 41.35 & 40.88  & 0.9860 & 0.9815 & 0.9858  & 64.90 \\
AHDRNet\cite{ahdrnet} & 43.63 & 42.93 & 41.14  & 0.9900 & 0.9849 & 0.9702  & 64.61 \\
NHDRRNet\cite{NHDRRNet} & 42.41 & 42.97 & 41.43  & 0.9877 & 0.9855 & 0.9857  & 61.21 \\
CEN-HDR\cite{cen-hdr} & 43.05 & 43.24 & 40.53  & 0.9908 & 0.9821 & 0.9856  &  64.34\\
HDRGAN\cite{hdrgan} & 43.92 & 44.03 & 41.57  & 0.9905 & 0.9851 & 0.9865  & 65.45 \\
HDR-Transformer\cite{hdrtransformer} & 44.32 & 44.23 & 42.18  & 0.9916 & 0.9924 & 0.9884  & 66.03 \\
\textbf{SCTNet} (Ours) & \textbf{44.49} & \textbf{44.27} & \textbf{42.29}  & \textbf{0.9924} & \textbf{0.9927} & \textbf{0.9887} & \textbf{66.65} \\
\hline

\hline

\hline
\end{tabular*}
\end{center}
\vspace{-5mm}
\end{table*}

\noindent \textbf{Shallow Features Extraction}:
Given the input $I_i \in  \mathbb{R} ^{6 \times H \times W}$, we first extract shallow features $f_i$ with convolution. Unlike existing methods that compute alignment matrices between the shallow features, we directly merge them and feed the fused features into our attention blocks.

\noindent \textbf{Global Spatial Attention Block}: To model the spatial correlation between different frames, we adopt the window-based multi-head self-attention following \cite{liu2021swin,swinir,hdrtransformer}. Technically, the merged features are first fed into the patch partition for tokenization. Then, the sequence of tokens is fed into the spatial attention block, which contains layer norm (LN), window-based multi-head self-attention (W-MSA), and a point-wise multi-layer perceptron (MLP). For the $j^{th}$ layer, $j\in \{1, ..., L\}$, it takes the sequence $z_{j-1}$ as input, and outputs the new sequence $z_{j}$:
\begin{equation}
\begin{split}
&\hat{z}_j = W\textnormal{-}MSA(LN(z_{j-1})) + z_{j-1}; \\
&z_j = MLP(LN(\hat{z}_j)) + \hat{z}_j.
\end{split}
\end{equation}

\noindent \textbf{Semantic-Consistent Attention Block}:
As shown in Figure \ref{fig:network_arch}, we propose a semantic-consistent attention block (S-CAB) to interact between different frames. Specifically, taking the spatially refined feature $z_j$, we first split it into three sub-groups $z'_{ii}$, where each group is linked with shallow features of each EV input $f_i$ by a residual addition. We obtain the linked feature $z'_{i}$ by:

\begin{equation}
z'_{i} = z'_{ii} + f_i.
\end{equation}

Then, we apply the channel-wise multi-head cross-attention (C-MCA) between them. Technically, we compute the query from the reference-linked feature, while the keys and values are generated from others. For simplicity, we take $z'_{1}$ and $z'_{2}$ as examples. Let $q_2$ be the query from $z'_{2}$ and $(k_1; v_1)$ be the key and value generated from $z'_{1}$, the
output $z'_{12}$  of our C-MCA becomes:

\begin{equation}
   z'_{12} = C-MCA(q_2, k_1, v_1) = softmax(\frac{q_2 \times k_1^T}{\sqrt{d_k}})v_1,
\end{equation}
where $d_k$ is the scaling factor. Similarly, we can obtain the cross-calibrated feature $z'_{32}$ by applying C-MCA on $z'_{2}$ and $z'_{3}$. Finally, $z'_{12}$, $z'_{22}$, and $z'_{32}$ are merged and fed into an FFN to generate the final output of our S-CAB. The computational cost of our C-MCA is:
\begin{equation}
   \Omega(C-MCA) = C^2 \times N / 9, 
\end{equation}
where $C$ is the channel dimension and $N$ is the spatial resolution. By extending the inter key-query correlation to cross-frame key-query correlation, our S-CAB block suggests a natural way to aggregate multi-EV features. Moreover, our cross-attention is applied on the channel axis, which can deeply preserve and explore the semantic consistency between different frames. 

\noindent \textbf{Skip Connection}: After the attention blocks, we build a skip connection from the reference feature $f_2$. The skip connection is realized through a convolution. Finally, we use another convolution to generate the final HDR output.

\subsection{Loss function}\label{ssc:loss_function}
Our network is trained end-to-end. We first supervise our network with $\mathcal{L}_1$ loss function. Note that Kalantari~\etal~\cite{Kalantari2017} has shown that computing the loss in the HDR domain leads to the appearance of discoloration effects. To avoid this, we compute the $\mathcal{L}_1$ loss in the tone-mapped domain using the $\mu$-law function:
\begin{equation}\label{eq:mulaw}
T(H) = \frac{log(1+\mu H)}{log(1+\mu)}, \quad \mu=5000,
\end{equation}
where H is the linear HDR image and $\mu$ is the amount of compression. Our $l_1$ loss for HDR imaging becomes:
\begin{equation}
\label{eq:l1}
\mathcal{L}_1 =  \lVert T(H) - T(\hat{H})  \rVert_1,
\end{equation}
where $T$ is the $\mu$-law tone mapping function, $H$  and $\hat{H}$ are the prediction and the ground truth in the HDR domain.

Additionally, following \cite{hdrtransformer}, we use an auxiliary perceptual loss $\mathcal{L}_p$ for supervision. The perceptual loss measures the difference between the feature representations of the output image and the ground truth image at multiple layers of the pre-trained CNN, realized by computing the mean squared error between the feature maps at each layer. The motivation for such a loss is to assess the dissimilarity between the predicted HDR image and the actual ground truth image. Formally, we have our perceptual loss:

\begin{equation}
\mathcal{L}_p = \sum_{j} \lVert \Phi(T(H)) - \Phi(T(\hat{H}))  \rVert_1,
\end{equation}
where $\Phi$ are the feature maps extracted from a pre-trained VGG-16 network with $j$ indicating the $j$-th layer. Finally, the training loss can be formulated as:

\begin{equation}
\label{eq:lossfunction}
\mathcal{L}=  \mathcal{L}_1(T, T_{GT}) + \alpha \cdot \mathcal{L}_p(T, T_{GT}),
\end{equation}
where $\alpha$ is a hyper-parameter set to 0.01.


\begin{figure*}
\begin{center}
\includegraphics[width=\linewidth,keepaspectratio]{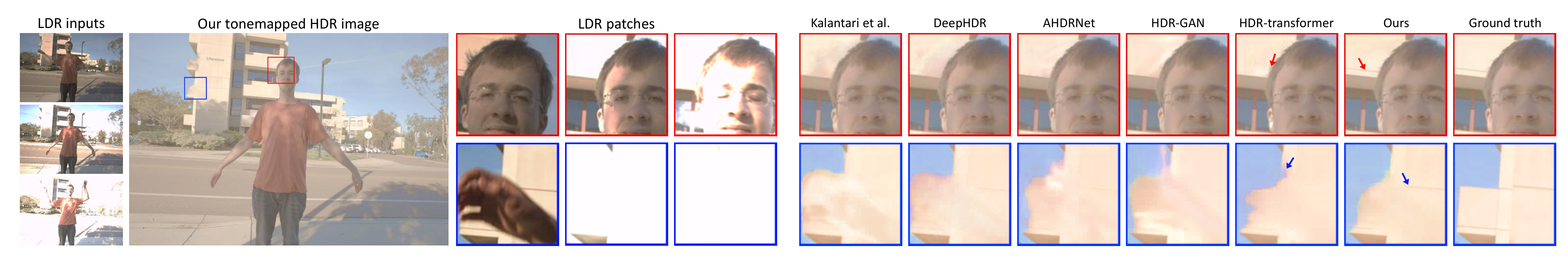}
\end{center}
\vspace{-6mm}
   \caption{Comparison of our proposed network with state-of-the-art methods on a Kalantari dataset's sample \cite{Kalantari2017}.    Thanks to the red patches, we show that our method is the most efficient to recover the ghost from the foreground motion in a bright area of the scene. With blue patches, we exhibit the ability of our solution to recover faithful information in occluded regions.
   Additionally, as demonstrated in both red and blue patches, our method excels in retrieving the building's texture information. Please zoom in for more details.}
\label{fig:result_kalantari}
\vspace{-1mm}

\end{figure*}

\section{Experiments}
\label{sec:experiments}
\subsection{Implementation Details}
\label{ssc:implementation_details}

\noindent \textbf{Datasets}:
We first evaluate our model on the conventional  Kalantari~\cite{Kalantari2017} benchmark composed of 74 training scenes and 15 testing scenes. We follow the common practice of cropping patches of size $128 \times 128$ with a stride of 64 from the training set. We use rotation and flipping as data augmentation. We also evaluate our proposed dataset which is composed of 108 training samples and 36 testing samples. Each sample represents the acquisition of a dynamic scene caused by large foreground or camera motions. The same dataset preparation and augmentation methods are applied.

\noindent \textbf{Implementation overview}: We implement our SCTNet on PyTorch with ADAM optimizer with an initial learning rate of 2e-4, $\beta_1$ to 0.9, $\beta_2$ to 0.999, and $\epsilon$ to 1e-8. We train the network on 2 V100 GPUs. All the counterparts' results are trained or downloaded from the official resources.

\noindent \textbf{Metrics}:
We use common metrics such as PSNR and SSIM, as well as HDR-VDP2 metric which was specifically designed for the HDR task. We compute PSNR and SSIM evaluation in linear and tone-mapped domains, denoted as $l-$ and $\mu-$, respectively. We also compute PU-PSNR/SSIM in the tone-mapped domain as proposed in \cite{pu21}. We set the display peak luminance to 100cd/m², a 1000:1 contrast, and an ambient light to 10~lux.

\subsection{Evaluation and Comparison }
\noindent \textbf{Evaluation on Kalantari~\etal Dataset}: We present the quantitative comparison in Table~\ref{table:quant}. We compare our method with 9 widely-used HDR methods, varying from early hand-crafted, lightweight, CNN, and Transformer methods. Specifically,  Hu \etal~\cite{Hu} and Sen \etal~\cite{PatchBasedHDR} are based on input patch registration methods. DHDRNet~\cite{Kalantari2017} is one of the pioneering learning-based methods that leverage the optical flow to align the inputs. Similarly, DeepHDR~\cite{deephdr} aligns the background feature by the homography and then feeds the aligned inputs into the classical encoder-decoder architecture. NHDRRNet~\cite{NHDRRNet} further leverages the non-local attention module during the learning stage. AHDRNet~\cite{ahdrnet} is the first method that introduces an attention block for feature alignment at the shallow stage. Then, the calibrated features are merged and fed into multiple dilated residual dense blocks for feature encoding.  HDRGAN~\cite{hdrgan} is the first GAN-based approach for HDR merging with a deep supervised HDR method, while HDR-Transformer~\cite{hdrtransformer} is the first work by replacing the conventional convolution operations by the transformer architecture. It can be seen that, with the standardized evaluation protocol, our method performs favorably against all the counterparts, setting new records on the conventional 
Kalantari~\etal~\cite{Kalantari2017}  dataset. A qualitative comparison can be found in Figure \ref{fig:result_kalantari}. It can be seen that our method can lead to prediction closer to the ground truth.


\begin{table}[t]
\small
\setlength\tabcolsep{0.6pt}
\renewcommand{\arraystretch}{1.0}
\begin{center}
\caption{Quantitative comparison on our dataset. All the models are trained through official implementations. Our method significantly outperforms all counterparts by a large margin.}
\label{table:quantour}
\begin{tabular}{m{2.4cm} | m{1cm} m{.9cm} m{1cm} m{.9cm}  m{.9cm}}
\hline

\hline

\hline
 Method & $\mu$-PSNR & $l$-PSNR& $\mu$-SSIM & $l$-SSIM &HDR-VDP2 \\
\noalign{\smallskip}
\hline
NHDRRNet \cite{NHDRRNet}  & 36.68 & 39.61  & 0.9590 & 0.9853 & 65.41 \\
DHDRNet \cite{Kalantari2017} & 40.05  & 43.37  & 0.9794  & 0.9924 & 67.09 \\
AHDRNet \cite{ahdrnet} & 42.08 & 45.30  & 0.9837  & 0.9943 & 68.80 \\
HDR-Transf. \cite{hdrtransformer} & 42.39 & 46.35  & 0.9844  & 0.9948 & 69.23\\
\textbf{SCTNet} (Ours) & \textbf{42.55}  & \textbf{47.51}  & \textbf{0.9850} & \textbf{0.9952}  & \textbf{70.66}  \\

\hline

\hline

\hline
\end{tabular}
\end{center}
\vspace{-3mm}
\end{table}

\begin{figure}
\begin{center}
\includegraphics[width=\linewidth,keepaspectratio]{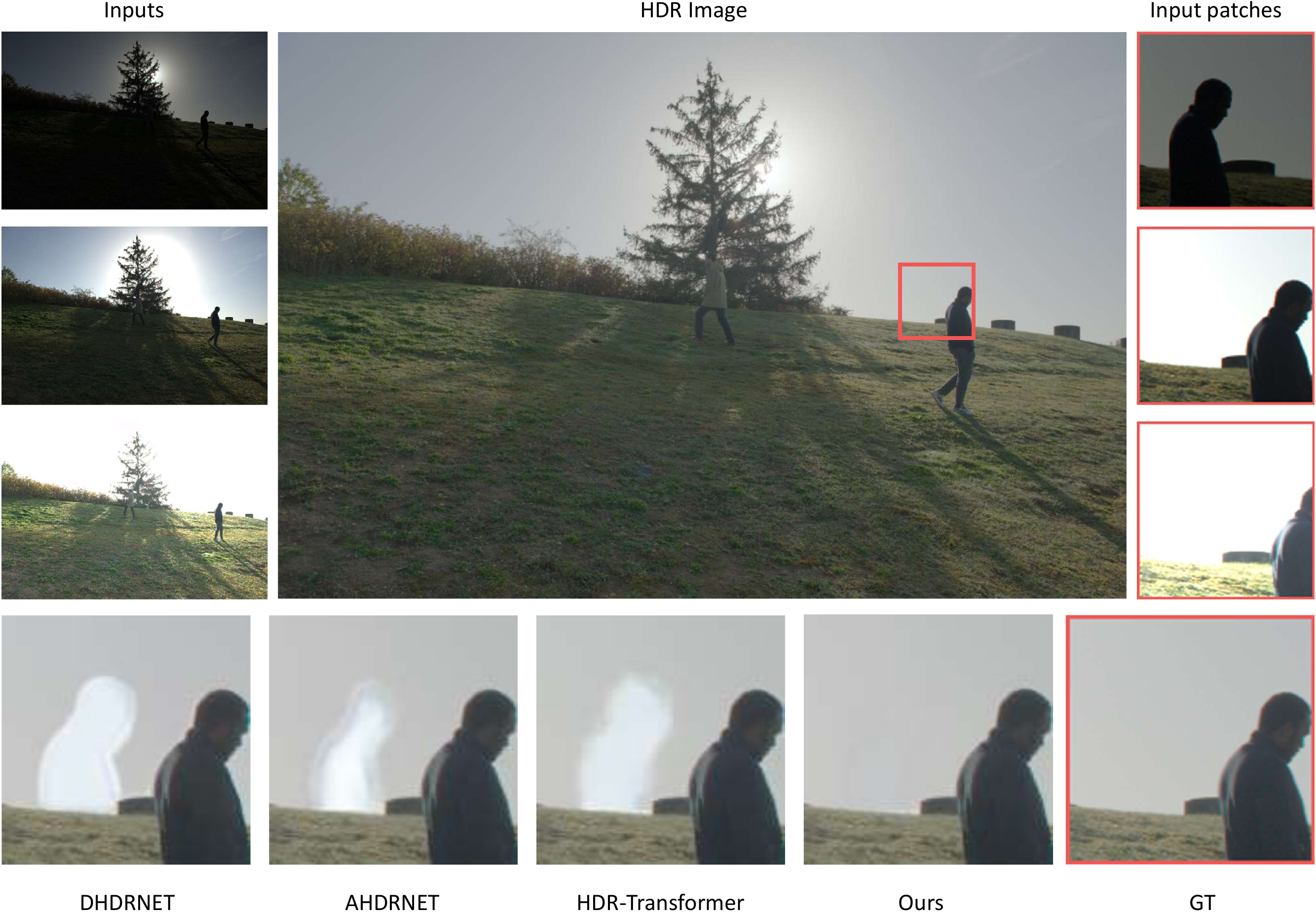}
\end{center}
\vspace{-4mm}
   \caption{Qualitative comparison on our proposed dataset. All the methods are trained using our training set. Our network outperforms all the SOTA counterparts in dealing with ghost artifacts due to the foreground motion. Please zoom in for details.}  
\label{fig:qual_our_dataset}
\vspace{-2mm}
\end{figure}

\begin{table}[t]
\small
\begin{center}
\setlength\tabcolsep{0.6pt}
\renewcommand{\arraystretch}{1.0}
\caption{Computational cost comparison of the proposed SCTNet solution against SOTA methods. The number of operations and parameters are evaluated using the script provided by the NTIRE \cite{ntire22} challenge. The input size is set to $1000 \times 1500$ pixels. The speed is measured on a Quadro P6000 GPU.}
\label{table:time}
\begin{tabular}{l|c c c c}
\hline

\hline

\hline
Method  & Params. (M) & GMACs & Size (Mb) & Speed (s)\\
\noalign{\smallskip}
\hline
\noalign{\smallskip}
DeepHDR \cite{deephdr}& 16.61 & 1453.70  & 66.5 &0.32\\
AHDRNet \cite{ahdrnet}& 1.52 & 2166.69  & \textbf{6.1} & 0.30\\
HDR-GAN \cite{hdrgan} & 2.63 & 778.81  & 10.6 &\textbf{0.29}\\
HDR-Transf. \cite{hdrtransformer} & 1.22 & 981.81  & 53.4 &8.65\\
\textbf{SCTNet} (Ours)  & \textbf{0.99} & \textbf{293.77}  & 29.1 &7.14\\

\hline

\hline

\hline
\end{tabular}
\end{center}
\vspace{-8mm}
\end{table}

\noindent \textbf{Evaluation on Our Dataset }: We also conduct comparisons with the most representative HDR methods on our proposed dataset. In Table~\ref{table:quantour}, it can be seen that our method sets new records on almost all the metrics, approving our versatility and effectiveness across different datasets. We provide qualitative results of our method in Figure \ref{fig:qual_our_dataset}. It can be seen that our method can better deal with object movement and generate the HDR output closer to the ground truth. This can be attributed to our attention blocks focusing on inter-image consistency.

\begin{figure*}
\begin{center}
\includegraphics[width=\linewidth,keepaspectratio]{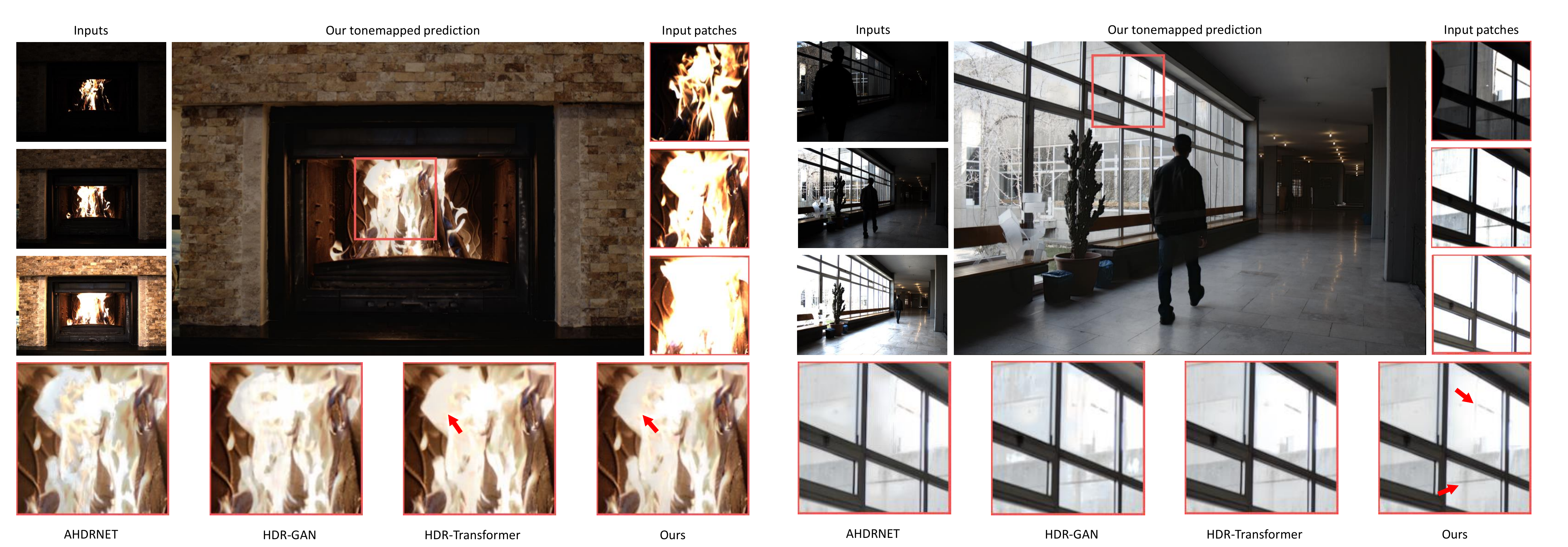}
\end{center}
\vspace{-5mm}
   \caption{We analyze the generalizability of our solution by evaluating on the unsupervised dataset proposed by Sen \etal \cite{tursun}. For a fair comparison, all networks are trained using our proposed dataset. In the first comparison (left), our network shows superior ability in restoring the illumination gradient resulting from the fire motion. In the second comparison (right), our method outperforms all the counterparts in recovering the maximum amount of information from the bright areas of the scene. Please zoom in for more details.}  
\label{fig:tursun}
\end{figure*}

\noindent \textbf{Generalization Capability}: We provide in Figure \ref{fig:tursun} the visual comparison on two unseen scenarios. We can observe that our method achieves consistently superior performance compared to all SOTA counterparts. Firstly we lead to better performance in restoring the illumination gradient resulting from the fire motion as shown in Figure\ref{fig:tursun}-Left. While dealing with foreground motion as shown in Figure\ref{fig:tursun}-Right, Our method outperforms others in recovering the maximum amount of information from the bright areas of the scene. These observations further validate our effectiveness.

\noindent \textbf{Comparison on the Computational Cost}: The computational cost of our method, as well as those of our counterparts, can be found in Table~\ref{table:time}. Our method achieves the best trade-off between the performance and the computational cost. It is worth noting that compared to the current SOTA HDR-Transformer~\cite{hdrtransformer}, our method only requires around 1/3 of the GMACs and 1/2 of the model size, validating our efficiency on attention blocks. Compared to AHDRNet, as shown in Table~\ref{table:quant} \& \ref{table:time} we achieve a notable improvement of +1dB on PSNR while using only 13\% of the GMACs and 65\% of the parameters, showcasing our efficiency.

\subsection{Ablation Studies}
\noindent \textbf{Key Component Analysis:} In this section, we conduct studies to verify the effectiveness and importance of our proposed components: G-SAB, S-CAB, and Skip connection to link with the shallow multi-EV features. The experimental results can be found in Table \ref{table:abla}. We gradually add our blocks on top of our baseline (denoted as Base.). It can be seen that each component brings improvement, validating the effectiveness of our proposed method. More ablation studies can be found in the supplementary material.

\begin{figure}[t]
\begin{center}
   \includegraphics[width=1\linewidth]{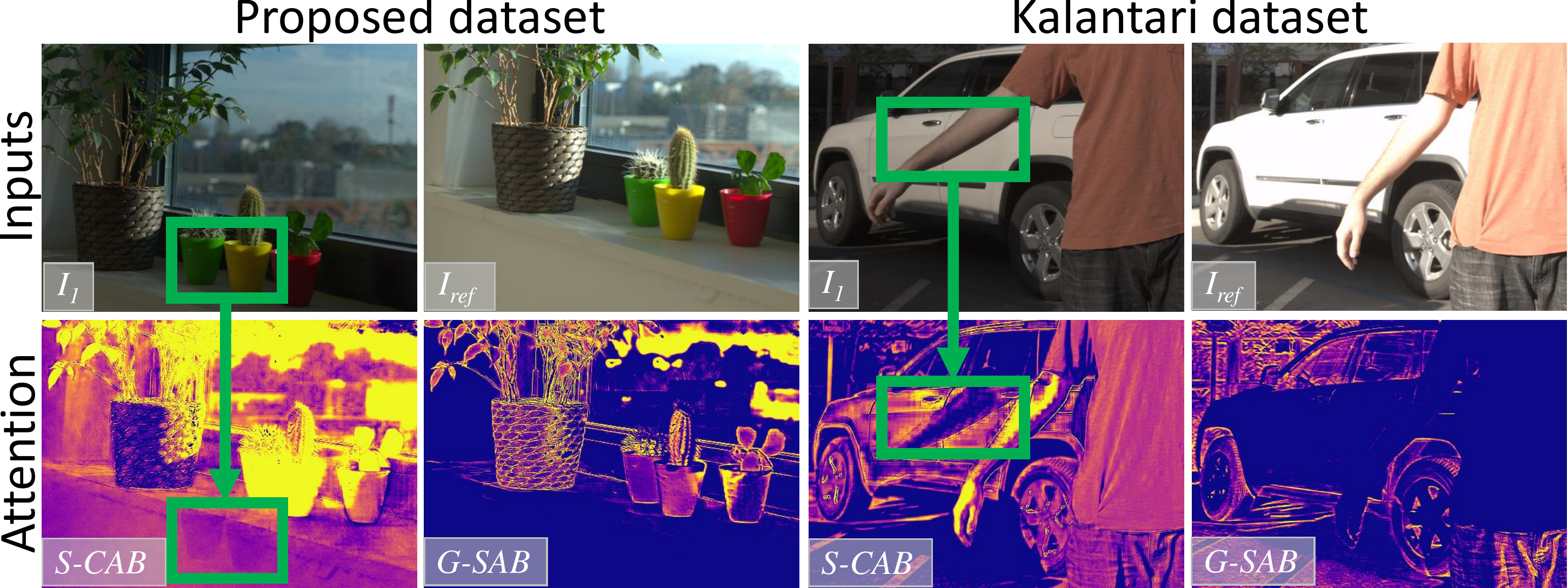}
\vspace{-5mm}
   \caption{$I_1$-$I_{ref}$ attention visualization.}
   \vspace{-5mm}
\label{fig:features}
\end{center}
\end{figure}

\noindent \textbf{Attention Visualization:} To better understand our proposed attention blocks, we provide feature visualizations between $I_1$ and the reference frame $I_{ref}$ in Figure \ref{fig:features}. It can be seen that: \textbf{(1)} Our G-SAB leverages global clues, such as shape and contour, while the S-CAB focuses on feature alignment across input frames. \textbf{(2)} The S-CAB leverages semantic clues to compute cross-image affinities, leading to activated attention on misaligned areas, as shown by the \textcolor{ForestGreen}{\textbf{green}} bounding boxes. \textbf{(3)} The complementary and simultaneous operation of our blocks leads to improved feature modeling for effective deghosting.

\begin{table}[t]
\small
\setlength\tabcolsep{0.6pt}
\renewcommand{\arraystretch}{1.0}
\begin{center}
\caption{Ablation study on the key components. The comparison is conducted on the Kalantari~\etal dataset.}
\label{table:abla}

\begin{tabular}{m{.9cm} m{1cm} m{1cm} m{.7cm} | l m{1cm} m{1.2cm} l }
\hline

\hline

 Base. & G-SAB & S-CAB & Skip & $\mu$-PSNR & PSNR& $\mu$-SSIM & SSIM \\

 \hline
 \cmark &  &   & & 43.42 & 41.68  &  0.9882 & 0.9861 \\
  \cmark & \cmark & & & 44.04 &41.89 &  0.9921  &0.9887  \\
 \cmark & \cmark & \cmark &  & 44.12 & 41.71	&	0.9923 &	0.9882\\
 \cmark & \cmark & \cmark & \cmark &\textbf{44.49} & \textbf{42.29}  & \textbf{0.9924} & \textbf{0.9887} \\

\hline

\hline
\end{tabular}
\end{center}
\vspace{-5mm}
\end{table}

\section{Conclusion and Future Work}

In this paper, we present a novel alignment-free network with a Semantics Consistent Transformer for HDR deghosting. Different from existing methods that estimate the image alignment by attention matrix at the shallow stage, our method is unified in that we jointly model the foreground motion and static semantic context simultaneously. We propose two types of attention blocks with intra-/inter-image interaction to directly deal with the alignment issues in the transformer architecture. Furthermore, we propose a novel dataset with more realistic samples and more variations on both lighting conditions and scene motions. 

\noindent \textbf{Future work}. There are several directions. First, although our method sets new records on both the existing Kalantari benchmark and our dataset, there is room for further improvement, especially in further reducing the computational cost while maintaining the performance. A second direction can be on the domain adaptation. How to improve the performance on one dataset while being trained with the other is yet an open question.  We hope our dataset can contribute to the development of methods of better adaptation and generalization capability.

\noindent\textbf{Ackowledgement} The authors thank the anonymous reviewers and ACs for their tremendous efforts and helpful comments. This research is financed in part by the Conseil R\'egional de Bourgogne-Franche-Comt\'e and the Alexander von Humboldt Foundation.

\newpage
{\small
\balance
\bibliographystyle{ieee_fullname}
\bibliography{egbib}
}

\end{document}